\documentclass[11pt]{article}

\usepackage[preprint]{acl}
\usepackage{times}
\usepackage{latexsym}

\usepackage[utf8]{inputenc}
\usepackage[T1]{fontenc}
\usepackage{hyperref}
\usepackage{url}
\usepackage{booktabs}
\usepackage{amsmath,amssymb,amsfonts}

\makeatletter
\@ifpackageloaded{lineno}{%
\newcommand*\patchAmsMathEnvironmentForLineno[1]{%
  \expandafter\let\csname old#1\expandafter\endcsname\csname #1\endcsname
  \expandafter\let\csname oldend#1\expandafter\endcsname\csname end#1\endcsname
  \renewenvironment{#1}%
     {\linenomath\csname old#1\endcsname}%
     {\csname oldend#1\endcsname\endlinenomath}}%
\newcommand*\patchBothAmsMathEnvironmentsForLineno[1]{%
  \patchAmsMathEnvironmentForLineno{#1}%
  \patchAmsMathEnvironmentForLineno{#1*}}%
\AtBeginDocument{%
\patchBothAmsMathEnvironmentsForLineno{equation}%
\patchBothAmsMathEnvironmentsForLineno{align}%
\patchBothAmsMathEnvironmentsForLineno{flalign}%
\patchBothAmsMathEnvironmentsForLineno{alignat}%
\patchBothAmsMathEnvironmentsForLineno{gather}%
\patchBothAmsMathEnvironmentsForLineno{multline}%
}%
}{}%
\makeatother
\usepackage{nicefrac}
\usepackage{microtype}
\usepackage{xcolor}
\usepackage{graphicx}
\usepackage{multirow}
\usepackage{makecell}
\usepackage{subcaption}
\usepackage{algorithm}
\usepackage{algorithmic}
\usepackage{fontawesome5}

\hypersetup{
  colorlinks=true,
  linkcolor=blue!70!black,
  citecolor=green!50!black,
  urlcolor=blue!70!black
}

\usepackage{xspace}
\newcommand{\method}{ESR\xspace}
\newcommand{\methodfull}{Early Stopping Rollout}
\newcommand{\KL}{\mathrm{KL}}

\newcommand{\beatsteacher}{\textcolor{red!85!black}{$\bigstar$}}

\title{Less is More: Early Stopping Rollout for On-Policy Distillation}

\author{%
  Zhou Ziheng\textsuperscript{1,\,\faEnvelope}, Jiaqi Li\textsuperscript{2}, Huacong Tang\textsuperscript{1}, Ying Nian Wu\textsuperscript{1}, Demetri Terzopoulos\textsuperscript{1} \\[4pt]
  \textsuperscript{1}University of California, Los Angeles \quad \textsuperscript{2}Beijing Institute of General Artificial Intelligence \\
  \faEnvelope\ \texttt{josephziheng@ucla.edu}
}

\begin{document}

\maketitle

\begin{abstract}
On-policy distillation has recently emerged as a promising alternative to standard sequence-level imitation, training a student by scoring its own rollouts with a teacher model. However, we observe ``Off-policy Teacher Decay'' problem in this paradigm: for the later tokens, with student's earlier trajectory as context that is off-policy to the teacher, the teacher's ability to produce a corrective score would decay, and may fall back to token-completion behavior learned in the pre-training stage. We empirically verify this problem, and we propose a simple method \methodfull{} (\method{}) to fix it: simply restricting the rollout generation to the first $N$ response tokens. We show that \method{} both surpasses the full rollout OPD performance across model size, family, tasks and traning regime, and exhibit much higher GPU efficiency and training stability, especially under cross model family scenarios. We further investigate the mechanism behind this surprising performance and discovered ``Cascading Alignment'' and ``Sub-mode Commitment'' effect of \method{} that may explain why it works effectively and even sometimes exceeding the teacher model performance. Besides, we show that this position-based token selection strategy cannot be fully explainable by KL divergence and entropy signals.
\end{abstract}

\section{Introduction}
\label{sec:intro}

On-policy distillation (OPD) has emerged as a dominant paradigm for model distillation in industrial practice. The student generates its own rollouts $\tau$, which are then scored by the teacher: at each token, the teacher's probability $\pi_{teacher}(\tau_{student} \mid x_{prompt})$ serves as the soft target for the student~\citep{agarwal2024onpolicy, gu2024minillm}.  Viewed through an RL lens, OPD can be understood as using the teacher as a dense, token-level reward model that judges the student's own behavior on a given prompt~\citep{thinkingmachines2025onpolicy}. 

\begin{figure*}[!t]
\centering
\caption{\textbf{(Left) Off-policy Teacher Decay.} The teacher loses accuracy quickly as the student-generated rollout grows over a few hundred tokens. MATH-500, avg@4 ($n{=}4$, $t{=}0.7$). Teacher = Qwen3-1.7B; student = Qwen2.5-Math-1.5B. After $\sim$300 student tokens the teacher has effectively been dragged down to student-baseline performance. \textbf{(Right) Rollout length $N$ sweep on MATH-500.} LoRA, Qwen2.5-Math-1.5B $\rightarrow$ Qwen3-1.7B; best avg@4 across training steps. OPD and the undistilled baseline are shown as horizontal references. Performance saturates for $N \in [50, 200]$ and all beat OPD.}
\label{fig:motivation}\label{fig:n_sweep}
\begin{minipage}[b]{0.48\textwidth}
    \centering
    \includegraphics[width=\linewidth]{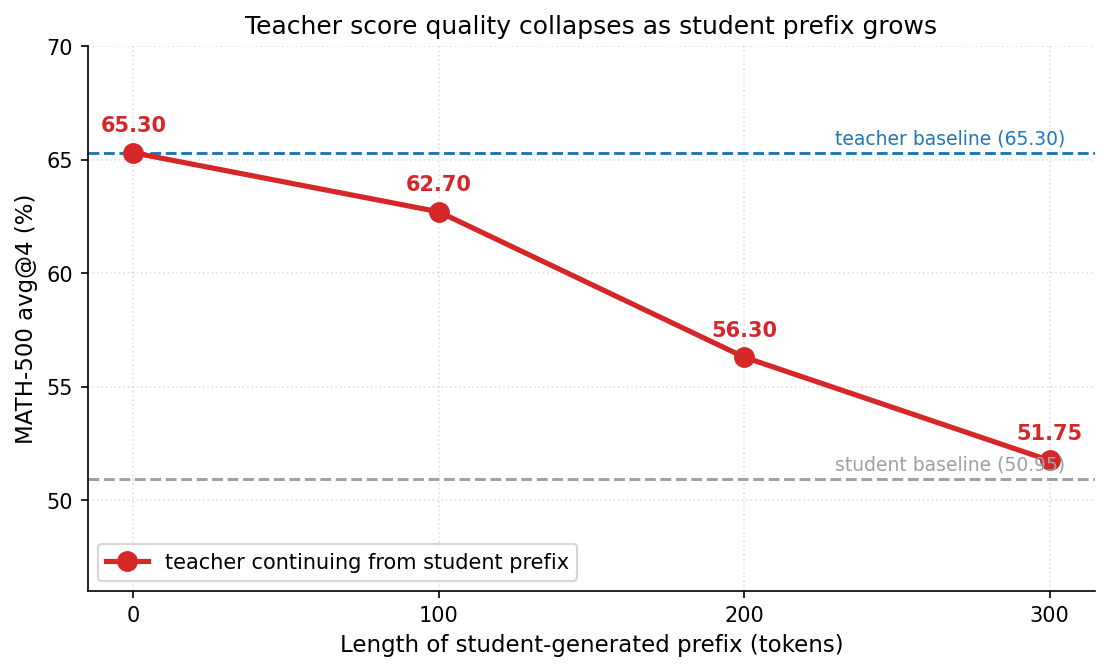}
\end{minipage}
\hfill
\begin{minipage}[b]{0.48\textwidth}
    \centering
    \includegraphics[width=\linewidth]{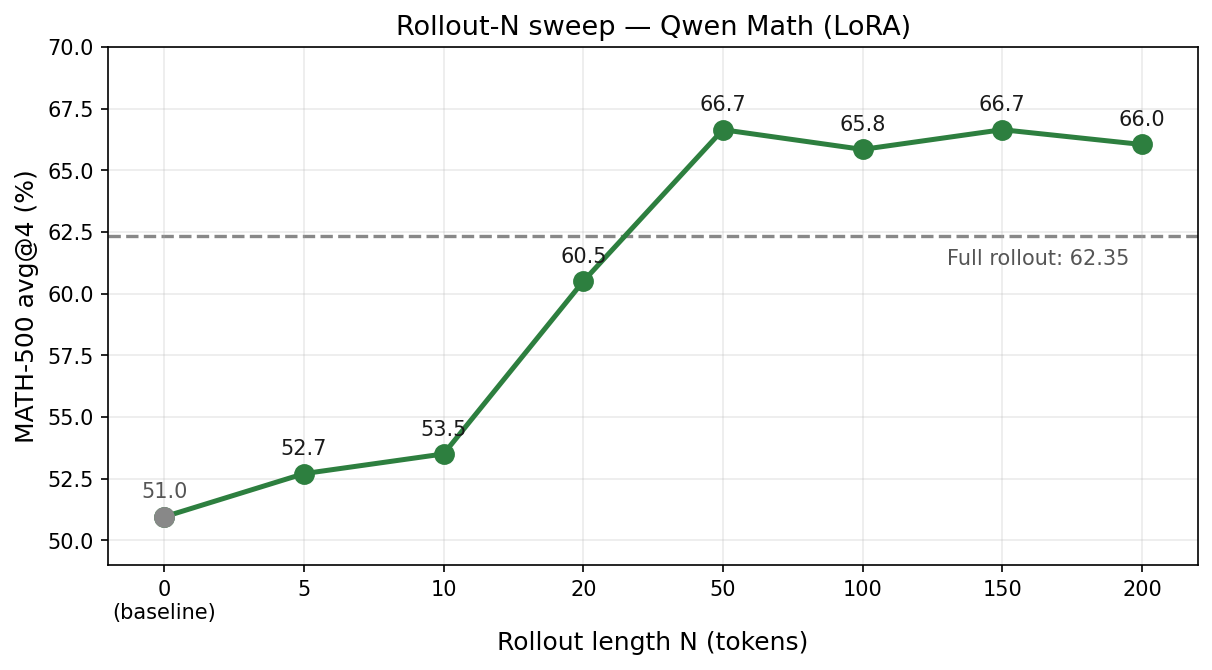}
\end{minipage}
\end{figure*}

However, we point out that the late-position token reward is ill-posed: at the first few tokens, the teacher's score is conditioned only on the prompt $\pi_{teacher}(\tau_{student}^{t=1} \mid x_{prompt})$ - indeed what we expect the teacher to score on. However, at a later position $m$, it becomes conditioned on the student's own previously generated tokens too: $\pi_{teacher}(\tau_{student}^{t=m} \mid x_{prompt}, \tau_{student}^{t=1:m})$. This conditioning context is off-policy to the teacher model, drifting away from teacher's model distribution. Recent works in the LLM alignment field find that LLMs may revert to pre-training behaviors when they see contexts not covered by their post-training~\citep{anthropic2025agenticmisalignment, tice2026alignmentpretraining, kutasov2026teachingclaudewhy}. Therefore, the teacher my no longer continue to \textit{correct} the student tokens to solve the answer but merely continues the auto-completion. We confirm and measure this decay by running a preliminary experiment by having the teacher to continue from an early-stopped student rollout. As shown in Figure~\ref{fig:motivation}, the teacher's performance decays toward the student's quickly after 100 tokens, and reaches the student baseline level within only 300 tokens. 

Motivated by this finding, we propose \methodfull (\method): restrict the student rollout to its first $N$ tokens and compute the distillation loss only on this early window. The change is a single line in any on-policy distillation loop. Despite its simplicity, \method consistently outperforms full-rollout OPD across tasks (math, code, function calling), training regimes (LoRA, full fine-tuning(FFT)), model scales (students 1.5B--32B, teachers 1.7B--72B), and model families (Qwen2.5, Qwen3, Gemma 2, Gemma 3), while reducing wall-clock cost by up to $24\times$ and peak training memory by up to $4\times$. Moreover, although normally the teacher is expected to be the upper bound of the distillation, we observe that \method-trained students can often \emph{exceed} the teacher.

Moreover, importantly, \method remains stable across model generations (eg. Qwen 2.5 to Qwen 3) and families (eg. Gemma to Qwen)(Table~\ref{tab:main_results}). We find that OPD brings little gain for same-family same-generation pairs, possibly due to that the teacher and student often share upstream data or were themselves co-distilled. The gain is much salient only when cross generation or cross family, but full-rollout OPD becomes very unstable in these settings and frequently collapses. Therefore, the stability and effectiveness of \method is very valuable.

To better understand the surprising effectiveness of \method, we conduct a series of ablation to investigate the potential reasons: 1) Firstly, we identified an important mechanism that we named as \textbf{Cascading Alignment} after training on the early window, KL divergence on the \emph{untrained} late tokens also drops by 30--40\%. Therefore, we find that with \method, the KL divergence does not have to see late positions to repair them. 2) Secondly, we discovered the \textbf{Sub-mode Commitment} behavior of \method that may explain why \method sometimes even \emph{exceeds} the teacher: the \method-trained student commits to a sub-mode of supported teacher modes instead of chasing the dominant mode. This sub-mode, however, may be better than the dominant mode sometimes. This finding indicates a potential path of superceding the teacher model in distillation that worth future investigations. 3) Lastly, we ablate over it \textbf{relevance to KL and entropy signals} and show that position is an independent factor from KL and entropy.

Our contributions are summarized below:
\begin{enumerate}
    \item \textbf{Method: \methodfull (\method) outperforms full-rollout on-policy distillation.} A one-line change---restricting the rollout length to the first $N$ response tokens---beats full rollout OPD distillation across tasks, model families, scales, and training, while being dramatically more efficient and stable to train, particularly for cross-family scenarios.
    \item \textbf{Deep dive: Investigation of why it works with systematic experiments.} We show with experiments that: 1) \method mitigates the Off-policy Teacher Decay from full-rollout OPD. 2) The Cascading Alignment effect enables \method to work for late-position tokens without training on them. 3) The Sub-mode Commitment behavior of \method enables it to even sometimes exceed the teacher.
\end{enumerate}

\section{Off-Policy Teacher Decay in OPD}
\label{sec:motivation}

We first identify a failure mode of on-policy distillation (OPD), which we call
\emph{Off-policy Teacher Decay}. In OPD, the teacher $T$ provides token-level
supervision by scoring the student $S$'s rollout at each position $t$, i.e.,
$\pi_T(\cdot \mid x, y^{S}_{<t})$, and the loss is typically averaged uniformly
across positions. This procedure implicitly assumes that, after conditioning on the
student's partial trajectory, the teacher can still provide a useful corrective
signal. However, as $t$ increases, the student
prefix $y^{S}_{<t}$, which is off-policy to the teacher, may move increasingly far away from the teacher's own
high-probability reasoning regions. The teacher may then no longer operating
from its natural reasoning state; instead, it could fall back to the behavior that completes next tokens from this
off-policy state induced by the student ~\citep{anthropic2025agenticmisalignment, tice2026alignmentpretraining, kutasov2026teachingclaudewhy}..

We propose that this drifting issue can be measured by the teacher's recoverability gap after conditioning on
a student-generated prefix:
\[
\Delta_{\mathrm{decay}}(t)
=
A_T(x)
-
A_T(x \mid y^{S}_{<t}),
\]
where $A_T(x)$ denotes the teacher's accuracy when solving from the original
prompt, and $A_T(x \mid y^{S}_{<t})$ denotes its accuracy when continuing from a
length-$t$ student-generated prefix. A larger $\Delta_{\mathrm{decay}}(t)$
indicates that the teacher is less able to recover from the student-induced
prefix, and therefore its late-position token distribution is less likely to
represent a reliable corrective target. 

To empirically verify this decay, we
feed the teacher a $k$-token student-generated prefix on MATH-500 and then let it
continue autoregressively. The teacher's avg@4 accuracy  decays from its unconditional baseline of $65.30\%$ to $62.70\%$ at
$N{=}100$, and further to $51.75\%$ at $N{=}300$, approaching the student-baseline
performance (Figure~\ref{fig:motivation}). This suggests that late-position
teacher scores are not independent assessments of the original problem; they
increasingly reflect how the teacher continues a trajectory that the student has
already committed to. Uniformly weighting all token positions in OPD gives undue emphasis over those regions where the teacher signal is no longer
corrective.

\section{Method: \methodfull{} (ESR)}
\label{sec:method}


Let $\pi_s$ denote the student and $\pi_t$ the teacher. In standard on-policy reverse-KL distillation, the student generates a response $\mathbf{y} = (y_1, \ldots, y_T)$ conditioned on prompt $x$, and the loss is
\begin{multline}
\mathcal{L}_{\text{full}} = \mathbb{E}_{\mathbf{y} \sim \pi_s(\cdot \mid x)} \Bigl[ \sum_{t=1}^{T} \KL\bigl(\pi_s(\cdot \mid x, \mathbf{y}_{<t}) \\
\,\|\, \pi_t(\cdot \mid x, \mathbf{y}_{<t})\bigr) \Bigr].
\label{eq:full_loss}
\end{multline}
\method{} (position cutoff $N$, with $N \ll T$ in practice) truncates the student rollout to its first $N$ tokens, and the loss is computed over exactly those tokens:
\begin{multline}
\mathcal{L}_{\text{ESR}}(N) = \mathbb{E}_{\mathbf{y} \sim \pi_s,\,|\mathbf{y}|\le N}\!\sum_{t=1}^{|\mathbf{y}|}\! \KL\bigl(\pi_s(\cdot \mid x, \mathbf{y}_{<t}) \\
\,\|\, \pi_t(\cdot \mid x, \mathbf{y}_{<t})\bigr).
\label{eq:esr_loss}
\end{multline}
If the student emits EOS before position $N$, the rollout terminates naturally. Everything else---generation temperature, LoRA target modules, optimizer, scorer---is unchanged from the standard on-policy KD loop.

\section{Main Experiments}
\label{sec:experiments}

\begin{table*}[!t]
\centering
\caption{\textbf{Main results: \method{} dominates \textbf{OPD} across same-family same-generation, same-family cross-generation, and cross-family pairs, and across scales (students 1.5B--32B, teachers 1.7B--72B) on MATH-500.} ``Student''/``Teacher'' columns are the base models with no distillation. OPD values are the peak across training; subscripts $^{\downarrow\!-\Delta}$ give the drop from peak to the final checkpoint, shown when >5\%. \emph{$^{\ddagger}$} values denote configurations that never reach a functional checkpoint (peak $<\!20$\% across all saved steps). \method{} uses $N{=}100$, LoRA. \textbf{Bold}: \method{} beats OPD. \beatsteacher: \method{} surpasses the teacher reference. For Gemma-2 2B $\to$ Qwen3-4B, \method{} uses $N{=}50$ on this pair. }
\label{tab:main_results}
\footnotesize
\setlength{\tabcolsep}{3pt}
\begin{tabular}{@{}l cccc | cccc@{}}
\toprule
& \multicolumn{4}{c|}{\textbf{avg@4}} & \multicolumn{4}{c}{\textbf{pass@4}} \\
\cmidrule(lr){2-5}\cmidrule(lr){6-9}
\textbf{Pair (Student $\to$ Teacher)} & \textbf{Student} & \textbf{Teacher} & \textbf{OPD} & \textbf{\method{}} & \textbf{Student} & \textbf{Teacher} & \textbf{OPD} & \textbf{\method{}} \\
\midrule
\multicolumn{9}{@{}l}{\textit{Same family, same generation}} \\
\midrule
Qwen3-1.7B $\to$ Qwen3-4B           & 69.20 & 77.95 & 65.85  & \textbf{69.20} & 81.00 & 86.40 & 78.00 & \textbf{81.20} \\
Qwen2.5-14B $\to$ Qwen2.5-Math-72B  & 73.80 & 72.60 & 73.45  & \textbf{74.30}\,\beatsteacher & 83.20 & 84.80 & 83.80 & \textbf{84.00} \\
Qwen2.5-32B $\to$ Qwen2.5-Math-72B  & 77.05 & 72.60 & 77.30\,\beatsteacher  & \textbf{78.10}\,\beatsteacher & 84.40 & 84.80 & 86.20\,\beatsteacher & \textbf{87.40}\,\beatsteacher \\
\specialrule{\heavyrulewidth}{2pt}{2pt}
\multicolumn{9}{@{}l}{\textit{Same family, cross generation}} \\
\midrule
Qwen2.5-Math-1.5B $\to$ Qwen3-1.7B  & 50.95 & 65.30 & 62.35 & \textbf{65.85}\,\beatsteacher & 72.80 & 77.00 & 75.20 & \textbf{79.80}\,\beatsteacher \\
Qwen2.5-Math-1.5B $\to$ Qwen3-4B    & 50.95 & 77.95 & 67.45$^{\downarrow\!-12.4}$ & \textbf{68.95} & 72.80 & 86.40 & 80.60 & \textbf{81.00} \\
Qwen2.5-Math-7B   $\to$ Qwen3-14B   & 53.60 & 76.15 & 68.85$^{\downarrow\!-6.5}$  & \textbf{68.95} & 75.00 & 83.20 & 80.00 & \textbf{81.20} \\
Qwen2.5-14B $\to$ Qwen3.5-35B-A3B   & 73.80 & 83.85 & 5.40$^{\ddagger}$  & \textbf{75.15} & 83.20 & 88.00 & 15.80$^{\ddagger}$ & \textbf{85.40} \\
Gemma-2 2B $\to$ Gemma-3 4B         & 13.45 & 66.60 & 22.95 & \textbf{27.20} & 28.20 & 74.80 & 31.40 & \textbf{39.40} \\
\specialrule{\heavyrulewidth}{2pt}{2pt}
\multicolumn{9}{@{}l}{\textit{Cross family}} \\
\midrule
Gemma-2 2B $\to$ Qwen3-4B        & 13.45 & 77.95 & 16.40$^{\downarrow\!-11.5}$ & \textbf{19.90} & 28.20 & 86.40 & 27.00$^{\downarrow\!-17.2}$ & \textbf{30.20} \\
\bottomrule
\end{tabular}
\end{table*}

\begin{table*}[t]
\caption{\textbf{(Left)} Performance on HumanEval with pass@1 and BFCL with full accuracy. \textbf{(Right)} Full Finetune (FFT) performance on MATH-500 with best across training steps. \textbf{Bold}: \method{} beats OPD. \beatsteacher: \method{} surpasses teacher. $^{\downarrow\!-\Delta}$ gives the drop from peak to final checkpoint (shown if >4\%).}
\label{tab:coding_bfcl}\label{tab:fullft_math}
\centering
\begin{minipage}[t]{0.48\textwidth}
\centering
\footnotesize
\setlength{\tabcolsep}{3pt}
\resizebox{\linewidth}{!}{%
\begin{tabular}{@{}l l cc@{}}
\toprule
\textbf{Pair} & \textbf{Method} & \textbf{HE} & \textbf{BFCL} \\
\midrule
\multirow{4}{*}{\shortstack[l]{Qwen2.5-Math-1.5B \\$\to$\,Qwen3-1.7B}}
 & Student    & 31.10 & 2.70 \\
 & Teacher  & 39.60 & 54.00 \\ & OPD                & 40.20$^{\downarrow\!-13.4}$ & 58.20\,\beatsteacher \\
 & \method{}          & \textbf{42.10}\,\beatsteacher & \textbf{61.30}\,\beatsteacher \\
\midrule
\multirow{4}{*}{\shortstack[l]{Gemma-2-2B\\$\to$\,Gemma-3 4B}}
 & Student   & 23.78 & 73.17 \\
 & Teacher  & 20.70 & 72.83 \\
 & OPD                & 22.00$^{\downarrow\!-10.4}$ & 76.83\,\beatsteacher \\
 & \method{}          & \textbf{28.70}\,\beatsteacher & \textbf{79.00}\,\beatsteacher \\
\bottomrule
\end{tabular}%
}
\end{minipage}%
\hfill
\begin{minipage}[t]{0.46\textwidth}
\centering
\footnotesize
\setlength{\tabcolsep}{3pt}
\resizebox{\linewidth}{!}{%
\begin{tabular}{@{}l l c c@{}}
\toprule
\textbf{Pair} & \textbf{Method} & \textbf{avg@4} & \textbf{pass@4} \\
\midrule
\multirow{4}{*}{\shortstack[l]{Qwen2.5-Math-1.5B \\$\to$\,Qwen3-1.7B}}
& Student  & 50.95  & 72.80  \\
& Teacher & 65.30  & 77.00 \\
 & OPD & 58.20 & 75.40 \\
 & \method{} & 56.20 & 73.80 \\
\midrule
\multirow{4}{*}{\shortstack[l]{Gemma-2-2B\\$\to$\,Gemma-3 4B}}
& Student & 13.45 & 28.20 \\
& Teacher  &66.60 &74.80 \\
 & OPD & 13.90 & 25.00 \\
 & \method{} & \textbf{26.65} & \textbf{40.40} \\
\bottomrule
\end{tabular}%
}
\end{minipage}
\end{table*}
\subsection{Setup}\label{sec:sectup}

\textbf{Models.} We evaluate across three regimes: \emph{same-family same-generation} (e.g.\ Qwen2.5$\to$Qwen2.5, Qwen3$\to$Qwen3), \emph{same-family cross-generation} (e.g.\ Qwen2.5$\to$Qwen3, Gemma-2$\to$Gemma-3), and \emph{cross-family} (Gemma$\to$Qwen). Student sizes range from 1.5B to 32B and teacher sizes from 1.7B to 72B.

\textbf{Training.} We employed reverse KL divergence loss with learning rate $5{\times}10^{-5}$), and generate sequences with temperature 0.7. Since we have many experiments, due to resource constraints, the main experiments use LoRA~\citep{hu2022lora} ($r{=}32$, $\alpha{=}64$. But we conduct full finetune ablations to confirm its validity. Each training step processes a batch of 16 problems with 1 rollout per problem ($n_{\text{samples}}{=}1$, batch size 16). We train for 200 steps on all tasks, and saving checkpoints every 50 steps. Training data are drawn from from NuminaMath~\citep{numinamath}, CodeUltraFeedback~\citep{weyssow2024codeultrafeedback}, and glaive-function-calling-v2~\citep{glaive2023functioncalling}. Our method uses $N{=}100$ unless otherwise specified. For pairs whose student and teacher use different tokenizers (all cross-generation and cross-family pairs in our setup), we decode the student rollout to text and re-encode it under the teacher's tokenizer to obtain teacher token-level log-probabilities; the reverse-KL loss is then computed on tokens that are token-aligned across the two vocabularies via a greedy text-span match.

\textbf{Evaluation.} MATH-500~\citep{hendrycks2021measuring, lightman2023verify} with $n{=}4$ samples at temperature 0.7 (reporting avg@4), HumanEval~\citep{chen2021evaluating, liu2024evalplus} at temperature 0.0 (reporting pass@1), BFCL~\citep{yan2024bfcl} reporting full accuracy: correct function name and arguments.

\subsection{Overall performance}
\label{sec:main_results_sub}

\paragraph{\boldmath\method beats OPD across model families, generations, sizes.} Across every cell of Table~\ref{tab:main_results}, \method{} matches or beats OPD's best score, and surpasses the teacher reference in many of them. For same family same generation setting, we test three sizes of model (Qwen 1.7B, 14B and 32B). Full rollout OPD sometimes even fall below its original performance (1.7B and 14B), but \method{} always improves. For cross generation setting, we test Qwen 2.5 - Qwen 3 or 3.5, with sizes ranging from 1.5B to 14B. We also tested Gemma 2 to 3 to ensure it works in different model series. For cross family settings, we let Gemma 2 2B to learn from Qwen3 4B. \method{} consistently exceed the full rollout training, with full rollout training collapse in most of the times.

\paragraph{\boldmath\method matches or beats OPD across tasks and training regimes (LoRA vs FFT.}  We tests in both Qwen series and Gemma series for task and training regime generalization. Table~\ref{tab:coding_bfcl} shows that \method{} is also better in coding (Human Evaluation, HE) and tool calling tasks(BFCL). Table~\ref{tab:fullft_math} reports FFT on MATH-500 for the Qwen2.5$\to$Qwen3 and Gemma-2$\to$Gemma-3 pairs. On Qwen \textbf{OPD} $58.20$ is slightly better than \method{} $56.20$ avg@4, but the gap is close. For Gemma \method{} dominates \textbf{OPD} by 12.75\% avg@4 and +15.40\% pass@4. \method{} is therefore the safer choice in both parameter regimes.

\subsection{Stability of \method{}}
\label{sec:stability}

\paragraph{\boldmath\method  is significantly more robust than OPD in training.} In cross-generation and cross-family settings, full-rollout distillation degrades or completely collapses most of the times; \method{} degrades nowhere. We denote the cells with degrading or collapsing failure mode in Table~\ref{tab:main_results} and Table~\ref{tab:coding_bfcl} with $^{\downarrow\!-\Delta}$ and $^{\ddagger}$. However, we observe the student to benefit significantly in these setting, showing more than 10 \% improvement for avg accuracy many times, whereas bare improvement can be observed in the same family same generation distillation setting.

\paragraph{\boldmath\methodfull is not sensitive to the choice of $N$ except for cross-family setting.} A natural question naturally occurs - how to choose where to stop? Is it sensitive? We conducted a set of sweeping experiments in Figure~\ref{fig:n_sweep} sweeps $N$ on MATH-500 with Qwen2.5-Math-1.5B and Qwen3 1.7B, a cross-generation setting where full rollout OPD suffers from stability issue, and reveals a robust region: it reaches just as good performance starting from $N{=}50$ and remains stable to $N{=}200$. The method is not sensitive to the exact choice of $N$ within a certain region. But we do find that for the cross-family setting (Gemma-Qwen pair), it is sensitive that it is stable with 50 tokens but not 100 tokens. Therefore, the bigger gap between the teacher and student, the more sensitive it is for choice of $N$. This also validates our ``Off Policy Teacher Decay'' diagnosis of OPD - the bigger gap between the student and teacher model, the more off-policy the student trajectory prefix is to the teacher and the bigger decay it causes.

\subsection{Efficiency of \method{}}
\label{sec:efficiency}

\begin{table}[t]
\centering
\caption{\textbf{Training efficiency.} Single A6000 (48\,GB), bs${=}16$, student teacher Qwen3-1.7B. ESR uses $N{=}100$; memory values in GB. We report the average running time and GPU memory usage across student model generation, training and teacher scoring phases. Note that the real time usage can be larger if there isn't enough GPU to hold the student and teacher models together and requires model loading and unloading.}
\label{tab:timing}
\footnotesize
\setlength{\tabcolsep}{2.5pt}
\resizebox{\columnwidth}{!}{%
\begin{tabular}{@{}l l cccc@{}}
\toprule
\textbf{Metric} & \textbf{Method} & \textbf{\makecell{Teacher\\Scoring}} & \textbf{\makecell{Student\\Generation}} & \textbf{\makecell{Student\\Training}} & \textbf{Total} \\
\midrule
\multirow{3}{*}{\shortstack[l]{-step\\wall time}}
 & ESR & 1\,s & 5\,s & 2\,s & \textbf{8\,s} \\
 & OPD            & 7\,s & 180\,s & 7\,s & \textbf{194\,s} \\
 & Speedup        & 7$\times$ & 36$\times$ & 3.5$\times$ & \textbf{24$\times$} \\
\specialrule{\heavyrulewidth}{2pt}{2pt}
\multirow{3}{*}{\shortstack[l]{Peak GPU\\memory}}
 & ESR & 7.3\,G  & 7.2\,G & 9.6\,G  & \textbf{24.1\,G} \\
 & OPD            & 14.9\,G & 8.9\,G & 39.5\,G & \textbf{63.3\,G} \\
 & Savings        & 2.0$\times$ & 1.2$\times$ & 4.1$\times$ & \textbf{2.6$\times$} \\
\bottomrule
\end{tabular}%
}
\end{table}

Table~\ref{tab:timing} shows that \method{} achieves a 24$\times$ wall-clock speedup and reduces peak training memory by $\sim 4\times$. The dominant cost in OPD is autoregressive generation ($180$\,s/step for sequences averaging ${\sim}1000$ tokens); \method{} generates only $N{=}100$ tokens (5\,s/step). Note that with \method, all the student and teacher models can be put in one A6000 GPU comfortably. In our own practice, it saves a further big time overhead of model loading and unloading that we do not report here. 

\section{More Analysis on Why \method{} Works }
\label{sec:mechanisms_section}


\subsection{The \textit{Cascading Alignment Effect} of \method{}}
\label{sec:mechanisms}

\begin{figure*}[t]
\centering
\caption{\textbf{(Left)} Early tokens are high on student entropy, teacher entropy, and KL divergence simultaneously. \textbf{(Right) The convergence cascade.} Per-position KL between distilled student and teacher, before vs.\ after \method training. Yellow band: positions $[0, 100]$ that actually receive training loss. Blue band: the KL gap closed by training. Positions $100+$---which see no direct training signal---drop to the same KL as the trained region, confirming that alignment on the early window cascades through the autoregressive rollout.}
\label{fig:three_curves}\label{fig:cascade}
\begin{minipage}[b]{0.48\textwidth}
    \centering
    \includegraphics[width=\linewidth]{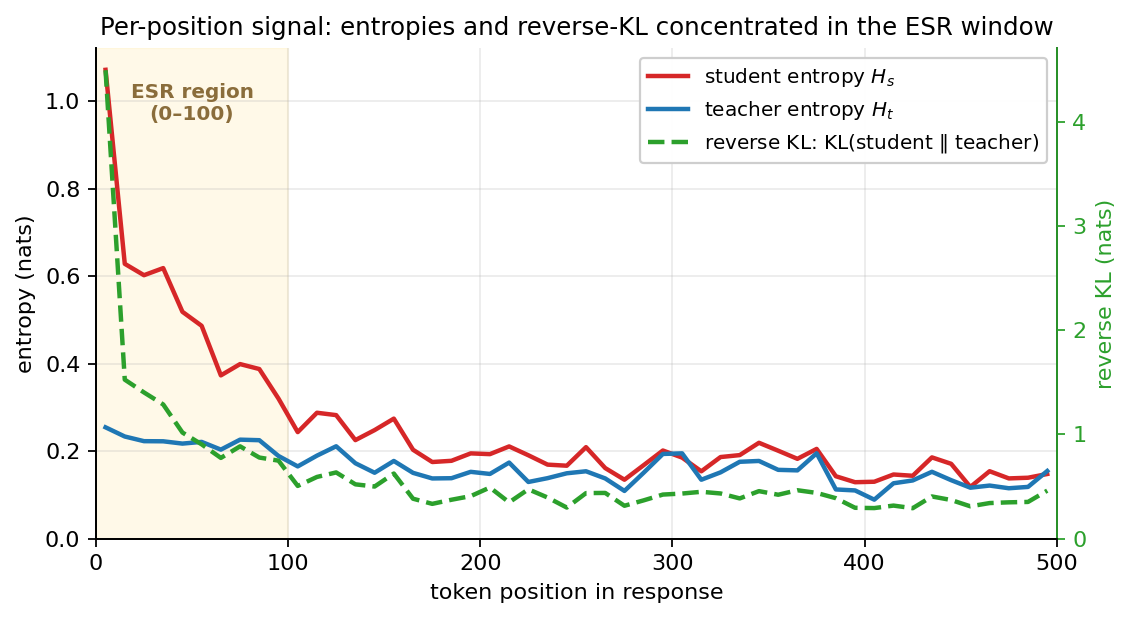}
\end{minipage}
\hfill
\begin{minipage}[b]{0.48\textwidth}
    \centering
    \includegraphics[width=\linewidth]{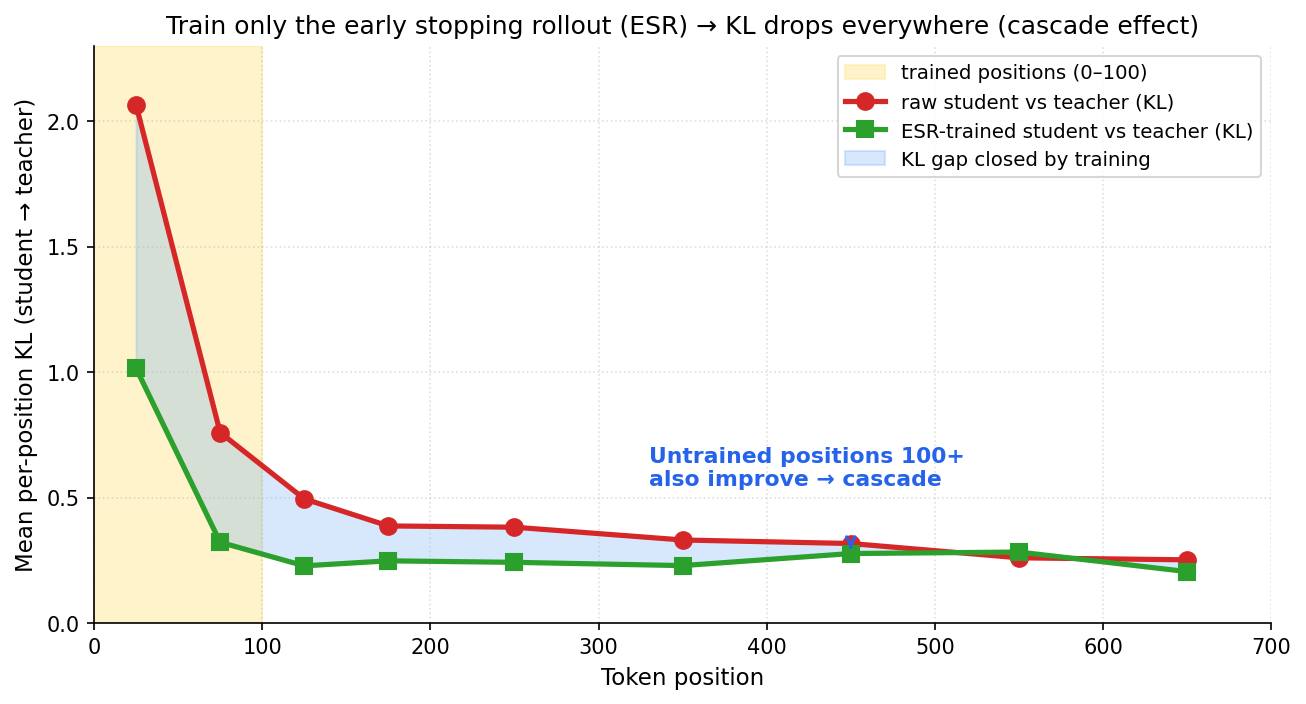}
\end{minipage}
\end{figure*}

Without training over the late-position tokens, can \method still learns the teacher behavior comprehensively? We find ``Convergence Cascade Effect'' of \method: even training on only the first $N$ tokens with \method, per-position KL divergence beyond $[0, N]$ region still drops by $30$--$40\%$ (Figure~\ref{fig:cascade}). This shows that the student can pick up the teacher's ``global mindset'' even with just the beginning tokens. 

Regarding to why \textit{Cascading Alignment Effect} happens, one reason that we suspect is that the beginning tokens often consist of problem framing and strategic planning content. The case study in Figure~\ref{fig:case_study} (Left) illustrates this concretely: on a representative MATH-500 trajectory, the first 100 tokens set up the geometry, name the unknown, and identify the key relationship (the altitude bisects the leg)---the choices that determine whether the rollout will succeed---while the last 100 tokens focus on executing the algebra that any solver can finish once the strategy is fixed. Therefore once the student picks up how to frame problems and plan the strategy, the later content naturally follows.

Moreover, recently \cite{sublinminal} shows student models may be able to learn the teacher's deep internal preference even with random numbers generated by the teacher, called ``subliminal learning''. Therefore, the early tokens may inject a global subliminal mindset to the student rather than only altering the prefix tokens.

\subsection{The \textit{Sub-mode Commitment Effect}}
\label{sec:mode_seek}

\method{} exceeds the teacher in many of the main experiments\ref{tab:main_results}. Even the full rollout OPD model slightly exceeds the teacher in function calling experiments a few times. This shows that student has the potential exceed the teacher even in normal OPD, and our method amplifies it. Why is so? Isn't teacher supposed to be the upper bound? 

\begin{figure*}[t]
    \centering
    \caption{\textbf{Mode-seeking on the planning region.} Schematic of reverse-KL behaviour at a multi-modal planning position. The teacher supports two modes (e.g.\ a verbose plan and a concise correct plan); reverse KL penalises student mass outside the support but not concentration within it. A short early-window loss can collapse the student onto the better supported mode, allowing the student to exceed the teacher's average behaviour. OPD training reverts the student toward the averaged teacher across late positions, undoing the concentration.}
    \label{fig:mode_seek}
    \includegraphics[width=\linewidth]{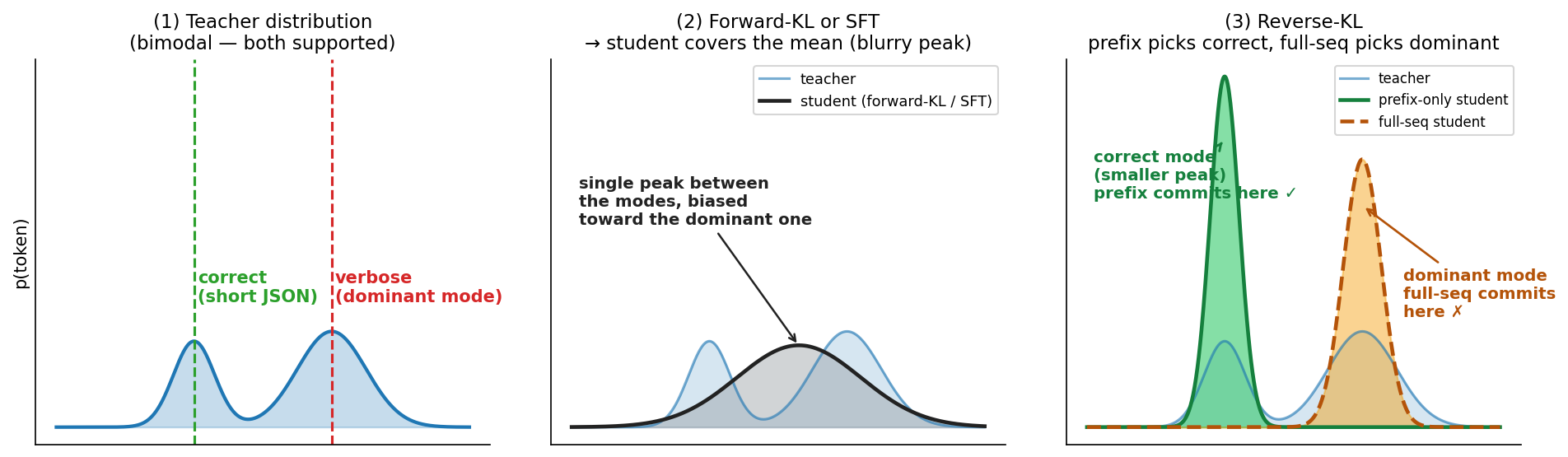}
\end{figure*}

We propose the reason lies in the mechanism of reverse KL $\mathrm{KL}(\pi_s \,\|\, \pi_t)$, which has a mode-seeking behavior: it penalizes the student for putting mass on tokens the teacher does not support, but not for concentrating mass on a single supported token. Therefore, the student has the possibility to land on a sub-mode of the teacher that is actually better. We visualize this mechanism in Figure~\ref{fig:mode_seek}, and verify it empirically below.

Indeed, we verified that in comparison to the full rollout OPD, \method can push the student more toward the non-dominant mode. We first scan the behavioral differences across distilled models. Surprisingly, \method-trained students produce sequences 2--3$\times$ shorter than the teacher, full-rollout, and even the base student itself: \method-100's median length is $\sim$380 tokens, against $\sim$1{,}150 for the teacher and $\sim$1{,}530 for full-rollout (Table~\ref{tab:behavioral_signature}, left). The teacher is substantially more verbose than the student, so distilling from such a teacher generally drags the student's length up --- and indeed full-rollout produces rollouts even longer than the teacher itself. The fact that \method{} learns from the \emph{same} teacher yet moves in the opposite direction shows how decisive the rollout-length choice is: by removing late-position supervision, the student preserves its own succinct style while still inheriting the teacher's reasoning strategy, leading to a more desirable outcome than simply copying the teacher.

Furthermore, we examine quantitatively how the student's probability output aligns with the teacher's modes. We take the top-10\% highest-KL tokens after training ($n{=}110{,}894$), which reveal behavioral differences most saliently, and check how often the student's top choice agrees with the teacher's top-1, falls in the teacher's top 2--5, or lies outside the top-5 (Table~\ref{tab:behavioral_signature}, right). We find that, indeed, \method{} produces a model that is more committed to the teacher's top 2--5 choices than to the teacher's top-1 ($47.4\%$ in top 2--5 vs $44.6\%$ for full-rollout; $41.9\%$ argmax agreement vs $45.7\%$). At the same time, \method's top-1 probability is higher than full-rollout's ($0.79$ vs $0.77$), showing the student is also more \emph{confident} in its own chosen token. This exactly shows that \method{} steers the student to commit to a secondary mode in the teacher's distribution.

\begin{table}[t]
\centering
\caption{\textbf{Behavioral Modes Analysis.}
  \textbf{(Left)} Response length distribution on -500. \method generates far shorter outputs than OPD, teacher, and its baseline, with 10th, 50th (median), 90th percentiles, and average length shown. 
  \textbf{(Right)} Alignment of student outputs with teacher distribution modes. Metrics include top-1 token probability (choice confidence), plus percentages of student top-1 matching teacher top-1 ($=$ top-1), falling in teacher top-2–5 ($\in$ top 2--5), or outside teacher top-5 ($\notin$ top-5). \method favors teacher non-top-1 choices and is more confident than OPD.
}
\label{tab:behavioral_signature}
\footnotesize
\setlength{\tabcolsep}{2.5pt}
\resizebox{\columnwidth}{!}{%
\begin{tabular}{@{}l l cccc@{}}
\toprule
& \textbf{Metric} & \textbf{\makecell{Student}} & \textbf{\makecell{Teacher}} & \textbf{OPD} & \textbf{\method} \\
\midrule
\multirow{5}{*}{\shortstack[l]{Response\\length\\(tokens)}}
 & 10\%  & $\sim$400     & $\sim$700     & $\sim$1{,}190 & $\sim$190 \\
 & 50\%  & $\sim$860     & $\sim$1{,}150 & $\sim$1{,}530 & $\sim$380 \\
 & 90\%  & $\sim$1{,}500 & $\sim$1{,}800 & $\sim$1{,}770 & $\sim$800 \\
 & Mean  & $\sim$990     & $\sim$1{,}210 & $\sim$1{,}480 & $\sim$460 \\
 & Ratio & $2.2\times$   & $2.6\times$   & $3.2\times$   & $1.0\times$ \\
\specialrule{\heavyrulewidth}{2pt}{2pt}
\multirow{4}{*}{\shortstack[l]{Student\\alignment\\to teacher\\probability}}
 & Top-1 prob       & 0.71   & ---  & 0.77            & \textbf{0.79} \\
 & $=$ top-1        & 28.6\% & ---  & \textbf{45.7\%} & 41.9\% \\
 & $\in$ top 2--5   & 59.8\% & ---  & 44.6\%          & \textbf{47.4\%} \\
 & $\notin$ top-5   & 11.5\% & ---  & 9.7\%           & \textbf{10.7\%} \\
\bottomrule
\end{tabular}%
}
\end{table}

\subsection{Ablation with KL and Entropy}
\label{sec:token_selection}

We find that, as shown in Figure~\ref{fig:three_curves}, early positions simultaneously have high KL divergence between student and teacher model, and high token entropy from both student and teacher models. This finding probes us to wonder if the effectiveness is intrinsically induced by the KL and entropy. To control these factors, we conduct a series of ablation experiments: pick the same amount (100) of tokens based on the highest KL divergence, highest student/teacher entropy or with them in combination with \method, regardless of position (Figure~\ref{tab:token_select}). If the effectiveness is indeed induced by the KL or entropy, then they should reproduce the same or better results. 

\begin{figure*}[t]
\centering
\caption{\textbf{(Left) Case study: early tokens encode strategy; later tokens mostly execute.} A representative MATH-500 trajectory. The first 100 tokens establish the geometric setup, name the unknown, and identify the key relationship (altitude bisects the leg)---these are the choices that determine whether the rollout will succeed. The last 100 tokens carry out an algebraic computation that any solver can finish once the strategy is fixed. \method{} supervises the first window; the second follows for free. \textbf{(Right) Token selection strategies on MATH-500.} Qwen2.5-Math-1.5B $\to$ Qwen3-1.7B, LoRA, $n_{\text{samples}}{=}1$, best avg@4 across training steps. All selectors pick $N{=}100$ tokens from full-length rollouts; \method{} is the only one that additionally truncates the rollout. Baseline (no distillation) 50.95\%; OPD 62.35\%. $H_s$: student entropy; $H_t$: teacher entropy; $\mathrm{RKL}$: reverse KL.}
\label{fig:case_study}\label{tab:token_select}
\begin{minipage}[b]{0.50\textwidth}
    \centering
    \fbox{\begin{minipage}{0.93\linewidth}
    \scriptsize
    \textbf{Problem.} In an isosceles right triangle, the altitude to the hypotenuse has length $4\sqrt{2}$. What is the area?

    \smallskip
    \textbf{First 100 tokens (strategy).} \emph{``Let's start by understanding the problem. In an isosceles right triangle, the two legs are equal, and the hypotenuse is $\sqrt{2}$ times the leg. If we denote each leg by $a$, then hypotenuse $= a\sqrt{2}$. The altitude to the hypotenuse is half the leg, because the altitude bisects the\dots''}

    \smallskip
    \textbf{Last 100 tokens (execution).} \emph{``\dots altitude from the right-angle vertex to the hypotenuse. This altitude has length $h = \tfrac{a \cdot a}{a\sqrt{2}} = \tfrac{a^{2}}{a\sqrt{2}} = \tfrac{a}{\sqrt{2}} = \tfrac{a\sqrt{2}}{2}$. Setting this equal to $4\sqrt{2}$: $\tfrac{a\sqrt{2}}{2} = 4\sqrt{2}$, so $a\sqrt{2} = 8\sqrt{2}$, giving $a = 8$. The area is $\tfrac{1}{2}\times 8\times 8 = \boxed{32}$.''}
    \end{minipage}}
\end{minipage}
\hfill
\begin{minipage}[b]{0.46\textwidth}
    \centering
    \small
    \begin{tabular}{lc}
    \toprule
    \textbf{Selection method} & \textbf{avg@4} \\
    \midrule
    ESR & \textbf{65.85} \\
    OPD & 62.35 \\
    \midrule
    Top-$\mathrm{RKL}$ & 53.35 \\
    Top-$H_t$ (teacher entropy) & 63.30 \\
    Top-$H_s$ (student entropy) & 62.70 \\
    $\mathrm{RKL} \!\cdot\! H_s$ & 56.90 \\
    $H_t \!\cdot\! H_s$ (product) & 55.35 \\
    $\mathrm{RKL} \!\cdot\! H_t \!\cdot\! H_s$ (triple product) & 57.90 \\
    \bottomrule
    \end{tabular}
\end{minipage}
\end{figure*}

 To our surprise, all underperform \method{}, and most of them also much underperform the OPD results. Teacher or student entropy based selection can match the full sequence by them alone, but their combination falls short significantly. What's also interesting is that KL divergence measure, the direct calculation of the loss magnitude, barely works. It only improves the baseline (50.95\%) for about 3 percent. And more surprisingly, we find that the largest 100 tokens of KL occupies around 93\% of the entire trajectory loss. This shows that the tokens that has larger signals are not necessarily the ones that have effective signals. 
 
 Therefore, although we don't exclude KL and entropy as potential mediator factor, we exclude them to be the sole factors that causes early tokens to be special. Position, therefore, should be considered an independent token selection dimension for the future.


\section{Related Work}
\label{sec:related_work}

\paragraph{Knowledge Distillation for Language Models.}
Knowledge distillation~\citep{hinton2015distilling} transfers knowledge from a teacher to a smaller student via soft targets, and \citet{kim2016sequence} extended this idea to sequence models with word-level and sequence-level objectives.
For autoregressive LLMs, both the divergence and the data distribution are crucial.
\citet{gu2024minillm} advocated reverse KL for generative LLM distillation, arguing that it avoids assigning mass to low-support teacher regions, and \citet{agarwal2024onpolicy} introduced Generalized Knowledge Distillation (GKD), which uses student-generated rollouts to obtain substantial gains over off-policy distillation on reasoning tasks.
Related work has explored other divergence and sampling choices, including skew KL and adaptive off-policy schedules~\citep{ko2024distillm}, general $f$-divergences~\citep{wen2023fdivergence}, the mode-seeking versus mean-seeking behavior of forward and reverse KL~\citep{wu2025rethinking}, and speculative knowledge distillation with interleaved teacher-student sampling~\citep{xu2025speculative}.
Our work builds directly on the on-policy reverse-KL setting of \citet{gu2024minillm} and \citet{agarwal2024onpolicy}, but asks a different question: holding the divergence and rollout distribution fixed, which token positions carry useful signal?

\paragraph{Token-Level Importance in Distillation and Reasoning.}
A growing line of work suggests that not all tokens contribute equally to learning.
In reasoning, \citet{wang2025beyond8020} found that only a small fraction of chain-of-thought tokens are high-entropy ``forking tokens'' that steer subsequent reasoning, while \citet{vassoyan2025ignorekl} showed that uniform KL penalties can suppress exploration on critical tokens and proposed entropy-weighted KL relaxation.
Related studies also identify token-level structure in planning and credit assignment, including preplan-and-anchor behavior~\citep{li2025preplan} and functional importance in reasoning chains~\citep{singh2026functional}.

In distillation specifically, several concurrent methods have explored token selection or weighting.
SelecTKD uses teacher verification to mask rejected tokens~\citep{huang2025selectkd}; AdaKD adapts token-level temperature based on training stability~\citep{xie2026adakd}; SE-KD disentangles selection along position, class, and sample axes and uses student-entropy filtering along the position axis~\citep{tavor2026rethinking}; and TSDKD combines entropy-based token selection with preference ranking~\citep{kim2026tsdkd}.
Our ablation in Section~\ref{sec:token_selection} shows that these scalar token-saliency criteria are insufficient: selecting the same number of tokens by top-KL, top-entropy, or combined entropy heuristics all underperform \method{}, and most underperform even plain full-sequence training.
This indicates that position is a load-bearing axis of supervision rather than a proxy for token saliency.
It also reconciles our findings with \citet{wang2025beyond8020}: in on-policy distillation, the high-entropy ``forking tokens'' are concentrated in the early uncontaminated window, so their conclusion is consistent with ours once position is taken into account.

\paragraph{Concurrent work.}
\citet{zhang2026prefix} independently report that concentrating OPD supervision on response prefixes is an effective efficiency lever; in their setting --- distilling a reasoning teacher into a base model that has not yet acquired reasoning behavior --- prefix OPD does not surpass full-trajectory OPD, suggesting that bootstrapping reasoning from scratch still benefits from full-trajectory supervision. Our setting is complementary: starting from a math-SFT student that already reasons, \method is able to achieve better performance than full rollout and may even push the student \emph{beyond} the teacher; we additionally provide systematic experiments and analyses on why late-position tokens are detrimental and on the mechanism through which prefix tokens drive learning --- none of which is addressed by this concurrent work.
\citet{li2026rethinkingopd} likewise note as a small part of their paper that the student's prefix can cause the teacher signal to degrade and that one may not use the full rollout. However, their focus is not on it alone, and the provided analysis and experiments are much narrower in scope than ours --- they do not run the systematic cross-generation, cross-family, cross-scale matrix used here, nor attribute the effect to specific mechanisms such as the convergence cascade or the reverse-KL mode-seeking behavior that lets \method-trained students surpass their teacher.

\section{Conclusion}
\label{sec:conclusion}

We introduce \methodfull{} (\method{}), a minimal one-line modification to OPD that constrains the rollouts to the first $N$ response tokens. Despite its simplicity, \method{} outperforms full-rollout OPD across three core dimensions: performance, efficiency, and stability across tasks, scales, and training regimes.


We also discovered a series of mechanisms that explains our method's efficacy. We discovered the ``Off Policy Teacher Decay'' as the root problem our method mitigates; the ``Cascading Alignment'' effect that may explain why it works effectively without training on the later tokens; and ``Sub-mode Commitment'' effect that explains why it even sometimes exceeds the teacher. Besides, we show that this position-based token selection strategy is an load-bearing axis beside KL divergence and entropy signals.

\paragraph{Limitations.} Our experiments expects the student models to be instruction-tuned model that already learns basic thinking ability rather than pre-trained only model, which may be inferior according to the concurrent work\citet{zhang2026prefix}. Our experiments are also focus on the setting where small open-source models (<100B) are finetuned for a specific task with a limited data budget. Whether the \method{} story holds at industrial scale general model capacity improvement (trillioin level model size; millions of training trajectories and above) remains unclear. It may be very likely that full rollout OPD works better in such level, although we still expect \method to be helpful under fixed budget setting. If one samples more trajectories to cover more diverse scenarios with shorter length, it is imaginable that it may be better than full rollout trajectories with narrower diversity if the ratio is calibrated well.  We also have not tested multi-modality or long-horizon tasks, which may exhibit different positional signal-quality patterns.


\paragraph{Ethical Considerations \& Potential Risks.} This work studies an algorithmic improvement to on-policy knowledge distillation; it does not target on subjective tasks like value alignment. We see no specific ethical and risk concerns beyond beyond those generally applicable to language-model training research.

\paragraph{Use of AI Assistants.} This paper was primarily conceived, designed, and drafted by the human authors. AI assistants (including ChatGPT and Claude) were used in a supporting role for proofreading, rewriting for clarity, and assisting with code development for the simulation platform. All scientific contributions, experimental design, analysis, and intellectual direction were driven by the authors, with AI tools serving as aids for language refinement and coding assistance.

\section*{Acknowledgments}
Funding and competing-interest disclosures will appear here in the final
version.

\bibliography{references_extended}

\clearpage
\appendix

\section{Training Efficiency}
\label{app:timing}

\begin{table}[!htbp]
\centering
\caption{\textbf{Training efficiency, detailed breakdown.} Per-step wall-clock time on a single A6000 (48GB). Student: Qwen2.5-Math-1.5B, LoRA, bs=16.}
\label{tab:timing_detail}
\small
\resizebox{\columnwidth}{!}{%
\begin{tabular}{llcccc}
\toprule
\textbf{Teacher} & \textbf{Method} & \textbf{Gen (s)} & \textbf{Score (s)} & \textbf{Train (s)} & \textbf{Total (s)} \\
\midrule
\multirow{2}{*}{Qwen3-1.7B} & OPD & 100--731 & 3--10 & 5--12 & ${\sim}$280 \\
 & Ours ($N{=}100$) & ${\sim}$5 & ${\sim}$1 & ${\sim}$2 & ${\sim}$8 \\
\midrule
\multirow{2}{*}{Qwen3-4B} & OPD & 97--733 & 3--9 & 6--10 & ${\sim}$210 \\
 & Ours ($N{=}100$) & ${\sim}$5 & ${\sim}$1 & ${\sim}$2 & ${\sim}$8 \\
\midrule
\multirow{2}{*}{Qwen3-8B} & OPD & 96--230 & 7--11 & 1--6 & ${\sim}$170$^\dagger$ \\
 & Ours ($N{=}100$) & ${\sim}$5 & ${\sim}$1 & ${\sim}$2 & ${\sim}$8 \\
\bottomrule
\multicolumn{6}{l}{\footnotesize $^\dagger$ Frequent OOMs; 48GB insufficient for 8B teacher + vLLM + student.}
\end{tabular}%
}
\end{table}


\section{Full Experimental Results}
\label{app:full_results}

\subsection{Math Results: Per-Step Performance}

Table~\ref{tab:app_math_main} presents per-step results for the primary math experiments (LoRA, $n{=}1$, 3,200 problems).

\begin{table}[!htbp]
\centering
\caption{\textbf{MATH-500 per-step results.} LoRA, $n{=}1$, $\text{bs}{=}16$, 3,200 problems. Baseline: 50.95\% avg@4.}
\label{tab:app_math_main}
\small
\resizebox{\columnwidth}{!}{%
\begin{tabular}{ll cccc}
\toprule
\textbf{Method} & \textbf{Metric} & \textbf{Step 50} & \textbf{Step 100} & \textbf{Step 150} & \textbf{Step 200} \\
\midrule
\multirow{3}{*}{ESR-50} & avg@4 & 62.35 & 66.05 & \textbf{66.65} & 64.85 \\
 & maj@4 & 69.40 & 72.00 & 71.00 & 71.20 \\
 & pass@4 & 77.20 & 79.40 & 81.00 & 79.60 \\
\midrule
\multirow{3}{*}{ESR-100} & avg@4 & 63.75 & 64.45 & 65.15 & \textbf{65.85} \\
 & maj@4 & 70.00 & 68.40 & 69.60 & 70.80 \\
 & pass@4 & 79.80 & 78.40 & 80.20 & 79.80 \\
\midrule
\multirow{3}{*}{ESR-150} & avg@4 & 65.35 & \textbf{66.65} & 65.30 & 65.75 \\
 & maj@4 & 66.80 & 67.00 & 66.30 & 67.30 \\
 & pass@4 & 79.00 & 81.00 & 78.20 & 80.00 \\
\midrule
\multirow{3}{*}{ESR-200} & avg@4 & \textbf{66.05} & 64.65 & 65.10 & 65.55 \\
 & maj@4 & 71.20 & 68.40 & 70.00 & 71.20 \\
 & pass@4 & 81.00 & 79.80 & 80.60 & 80.60 \\
\bottomrule
\end{tabular}%
}
\end{table}

\subsection{Math Results: Full Per-Step Trajectories ($n{=}1$, 3{,}200 problems, $\text{bs}{=}16$)}

\begin{table}[!htbp]
\centering
\caption{\textbf{Complete MATH-500 results, LoRA, $n{=}1$, $\text{bs}{=}16$, 3{,}200 problems.} Best per configuration in \textbf{bold}.}
\label{tab:app_math_n16}
\footnotesize
\resizebox{\columnwidth}{!}{%
\begin{tabular}{lcccc}
\toprule
\textbf{Config} & \textbf{Step 50} & \textbf{Step 100} & \textbf{Step 150} & \textbf{Step 200} \\
\midrule
\multicolumn{5}{l}{\textit{avg@4}} \\
\midrule
ESR-50    & 62.35 & 66.05 & \textbf{66.65} & 64.85 \\
ESR-100   & 63.75 & 64.45 & 65.15 & \textbf{65.85} \\
ESR-150   & 65.35 & \textbf{66.65} & 65.30 & 65.75 \\
ESR-200   & \textbf{66.05} & 64.65 & 65.10 & 65.55 \\
OPD  & 61.00 & 62.00 & \textbf{62.35} & 61.20 \\
\midrule
\multicolumn{5}{l}{\textit{pass@4}} \\
\midrule
ESR-50    & 77.20 & 79.40 & \textbf{81.00} & 79.60 \\
ESR-100   & 79.80 & 78.40 & \textbf{80.20} & 79.80 \\
ESR-150   & 79.00 & \textbf{81.00} & 78.20 & 80.00 \\
ESR-200   & \textbf{81.00} & 79.80 & 80.60 & 80.60 \\
OPD  & 74.60 & \textbf{75.20} & 74.60 & 75.00 \\
\bottomrule
\end{tabular}%
}
\end{table}

\subsection{Coding Results}

\begin{table}[!htbp]
\centering
\caption{\textbf{Complete coding results, LoRA.} HumanEval (HE) pass@1.}
\label{tab:app_coding_lora}
\footnotesize
\resizebox{\columnwidth}{!}{%
\begin{tabular}{lcccccccc}
\toprule
\textbf{Config} & \textbf{s50} & \textbf{s100} & \textbf{s150} & \textbf{s200} & \textbf{s250} & \textbf{s300} & \textbf{s350} & \textbf{s400} \\
\midrule
ESR-50    & 37.8 & 39.0 & 39.6 & 41.5 & 40.2 & 40.9 & \textbf{42.1} & 40.9 \\
ESR-100   & 37.2 & 39.0 & \textbf{42.1} & 37.8 & 39.0 & 37.8 & 37.8 & 38.4 \\
ESR-150   & 36.6 & 35.4 & 36.6 & 39.0 & \textbf{41.5} & 39.6 & 38.4 & 37.2 \\
OPD  & \textbf{40.2} & 31.7 & 32.3 & 32.9 & 27.4 & 28.0 & 26.8 & 26.8 \\
\bottomrule
\end{tabular}%
}
\end{table}

\subsection{Function Calling Results}

\begin{table}[!htbp]
\centering
\caption{\textbf{Function calling results (BFCL), LoRA.} Name accuracy / Full accuracy / Parse rate. Best full\_acc in \textbf{bold}.}
\label{tab:app_funcall}
\small
\resizebox{\columnwidth}{!}{%
\begin{tabular}{lccccc}
\toprule
\textbf{Method} & \textbf{Best Step} & \textbf{Name Acc} & \textbf{Full Acc} & \textbf{Parse Rate} \\
\midrule
Baseline & --- & 9.70\% & 2.70\% & 24.20\% \\
Teacher (Qwen3-1.7B) & --- & 75.30\% & 54.00\% & 75.30\% \\
\midrule
ESR-50 & 200 & 95.20\% & 57.20\% & 98.30\% \\
\textbf{ESR-100} & \textbf{100} & 86.20\% & \textbf{61.30\%} & 91.30\% \\
ESR-150 & 200 & 88.70\% & 61.50\% & 92.50\% \\
ESR-200 & 200 & 80.80\% & 54.50\% & 90.20\% \\
OPD & 100 & 81.00\% & 58.20\% & 86.70\% \\
\bottomrule
\end{tabular}%
}
\end{table}

\section{Token Classification Methodology}
\label{app:token_classification}

We classify each token into six categories based on string matching:
\begin{enumerate}
    \item \textbf{planning}: Reasoning keywords (``To'', ``Let'', ``First'', ``Step'', ``We'', ``Given'', ``Therefore'', ``Thus'', ``Since'').
    \item \textbf{structural}: Punctuation, whitespace, formatting tokens.
    \item \textbf{math\_number}: Digits (0--9).
    \item \textbf{math\_operator}: Arithmetic operators ($+$, $-$, $\times$, $/$, $=$).
    \item \textbf{math\_latex}: LaTeX delimiters (\texttt{\textbackslash(}, \texttt{\textbackslash[}).
    \item \textbf{continuation}: All others.
\end{enumerate}

\begin{table}[!htbp]
\centering
\caption{\textbf{Mean KL by token category and position range.}}
\label{tab:app_kl_by_category}
\small
\resizebox{\columnwidth}{!}{%
\begin{tabular}{lcccccc}
\toprule
\textbf{Category} & \textbf{0--4} & \textbf{5--19} & \textbf{20--49} & \textbf{50--99} & \textbf{100--199} & \textbf{200--499} \\
\midrule
planning & 4.50 & 0.79 & 1.49 & 1.66 & 1.49 & 2.37 \\
structural & 3.26 & 1.46 & 1.60 & 0.93 & 0.60 & 0.86 \\
math\_number & 1.49 & 0.60 & 0.74 & 0.28 & 0.17 & 0.13 \\
math\_operator & 7.30 & 1.84 & 0.81 & 0.37 & 0.21 & 0.14 \\
math\_latex & 8.84 & 9.23 & 6.50 & 4.95 & 2.97 & 1.87 \\
continuation & 1.94 & 1.12 & 1.19 & 0.89 & 0.70 & 0.48 \\
\bottomrule
\end{tabular}%
}
\end{table}

\begin{table}[!htbp]
\centering
\caption{\textbf{Top 20 highest-KL tokens} (minimum 50 occurrences across 10,000 trajectories).}
\label{tab:app_high_kl_tokens}
\small
\resizebox{\columnwidth}{!}{%
\begin{tabular}{clccc}
\toprule
\textbf{Rank} & \textbf{Token} & \textbf{Category} & \textbf{Count} & \textbf{Mean KL} \\
\midrule
1 & ``Solution'' & planning & 152 & 21.93 \\
2 & ``Analysis'' & continuation & 125 & 16.49 \\
3 & \texttt{\textbackslash[} & math\_latex & 7,152 & 13.21 \\
4 & ``examines'' & continuation & 74 & 11.51 \\
5 & ``He'' & continuation & 150 & 10.80 \\
6 & \texttt{\textbackslash(} & math\_latex & 21,243 & 10.30 \\
7 & ``First'' & planning & 1,706 & 9.98 \\
8 & ``tests'' & continuation & 52 & 8.71 \\
9 & \texttt{\textbackslash\textbackslash} & math\_latex & 82 & 8.68 \\
10 & ``There'' & continuation & 201 & 8.28 \\
11 & ``Therefore'' & planning & 4,913 & 7.95 \\
12 & ``Identify'' & continuation & 1,345 & 7.78 \\
13 & ``To'' & planning & 8,806 & 6.90 \\
14 & ``When'' & continuation & 89 & 6.62 \\
15 & ``This'' & continuation & 621 & 6.34 \\
16 & ``Thus'' & planning & 2,547 & 6.24 \\
17 & ``First'' (space) & planning & 259 & 6.10 \\
18 & ``To'' (space) & planning & 1,174 & 5.37 \\
19 & ``The'' & planning & 2,626 & 5.25 \\
20 & ``Next'' & planning & 1,753 & 5.03 \\
\bottomrule
\end{tabular}%
}
\end{table}

\section{Assets and Licenses}
\label{app:licenses}

Table~\ref{tab:assets_licenses} lists the models and datasets used in this paper, with their providers and licenses. All assets are used in accordance with their respective terms of use.

\begin{table}[h]
\centering
\caption{\textbf{Assets used in this paper.} All licenses verified at time of submission.}
\label{tab:assets_licenses}
\small
\resizebox{\columnwidth}{!}{%
\begin{tabular}{llll}
\toprule
\textbf{Asset} & \textbf{Type} & \textbf{Provider} & \textbf{License} \\
\midrule
Qwen2.5-Math-1.5B & Model & Alibaba & Apache 2.0 \\
Qwen2.5-Math-7B & Model & Alibaba & Apache 2.0 \\
Qwen3-1.7B / 4B / 8B / 14B & Model & Alibaba & Apache 2.0 \\
Gemma-2-2B & Model & Google & Gemma Terms of Use \\
Gemma-3-4B & Model & Google & Gemma Terms of Use \\
NuminaMath & Dataset & Numina & Apache 2.0 \\
CodeUltraFeedback & Dataset & Coseal & MIT \\
glaive-function-calling-v2 & Dataset & Glaive AI & Apache 2.0 \\
MATH-500 & Benchmark & Hendrycks et al. & MIT \\
HumanEval / HumanEval+ & Benchmark & OpenAI / EvalPlus & MIT \\
BFCL & Benchmark & UC Berkeley & Apache 2.0 \\
\bottomrule
\end{tabular}%
}
\end{table}

\clearpage
\newpage

\end{document}